%% file: main.tex
\documentclass[10pt,twocolumn,letterpaper]{article}

\usepackage{iccv}              


\definecolor{iccvblue}{rgb}{0.21,0.49,0.74}
\usepackage[pagebackref,breaklinks,colorlinks,allcolors=iccvblue]{hyperref}
\usepackage[table]{xcolor} 
\usepackage[accsupp]{axessibility} 

\def\paperID{*****} 
\def\confName{ICCV}
\def\confYear{2025}

\title{RG-Attn: Radian Glue Attention for Multi-modal Multi-agent Cooperative Perception}

\author{
Lantao Li\textsuperscript{1}, 
Kang Yang\textsuperscript{1,2}, 
Wenqi Zhang\textsuperscript{1}, 
Xiaoxue Wang\textsuperscript{1}, 
Chen Sun\textsuperscript{1}\\
\textsuperscript{1}Sony (China) Limited  \quad        \textsuperscript{2}Renmin University of China\\
{\tt\small \{lantao.li, wenqi.zhang, xiaoxue.wang, chen.sun\}@sony.com; yangkang1205@ruc.edu.cn}
}

\begin{document}
\maketitle
\input{sec/0_abstract}    
\input{sec/1_intro}
\input{sec/2_relatedworks}
\input{sec/3_methodology}

\input{sec/4_experiments}

\input{sec/5_conclusion}
{
    \small
    \bibliographystyle{ieeenat_fullname}
    \bibliography{main}
}

\end{document}

%% file: sec/0_abstract.tex
\begin{abstract}
Cooperative perception enhances autonomous driving by leveraging Vehicle-to-Everything (V2X) communication for multi-agent sensor fusion. However, most existing methods rely on single-modal data sharing, limiting fusion performance—particularly in heterogeneous sensor settings involving both LiDAR and cameras across vehicles and roadside units (RSUs). To address this, we propose Radian Glue Attention (RG-Attn)—a lightweight and generalizable cross-modal fusion module that unifies intra-agent and inter-agent fusion via transformation-based coordinate alignment and a unified sampling/inversion strategy. RG-Attn efficiently aligns features through a radian-based attention constraint, operating column-wise on geometrically consistent regions to reduce overhead and preserve spatial coherence, thereby enabling accurate and robust fusion. Building upon RG-Attn, we propose three cooperative architectures. The first, Paint-To-Puzzle (PTP), prioritizes communication efficiency but assumes all agents have LiDAR, optionally paired with cameras. The second, Co-Sketching-Co-Coloring (CoS-CoCo), offers maximal flexibility, supporting any sensor setup (e.g., LiDAR-only, camera-only, or both) and enabling strong cross-modal generalization for real-world deployment. The third, Pyramid-RG-Attn Fusion (PRGAF), aims for peak detection accuracy with the highest computational overhead. Extensive evaluations on simulated and real-world datasets show our framework delivers state-of-the-art detection accuracy with high flexibility and efficiency. \href{https://github.com/LantaoLi/RG-Attn}{GitHub Link}.
\end{abstract}

%% file: sec/1_intro.tex
\section{Introduction}
\label{sec:intro}

\begin{figure}[!t]
\centerline{\includegraphics[width=\columnwidth]{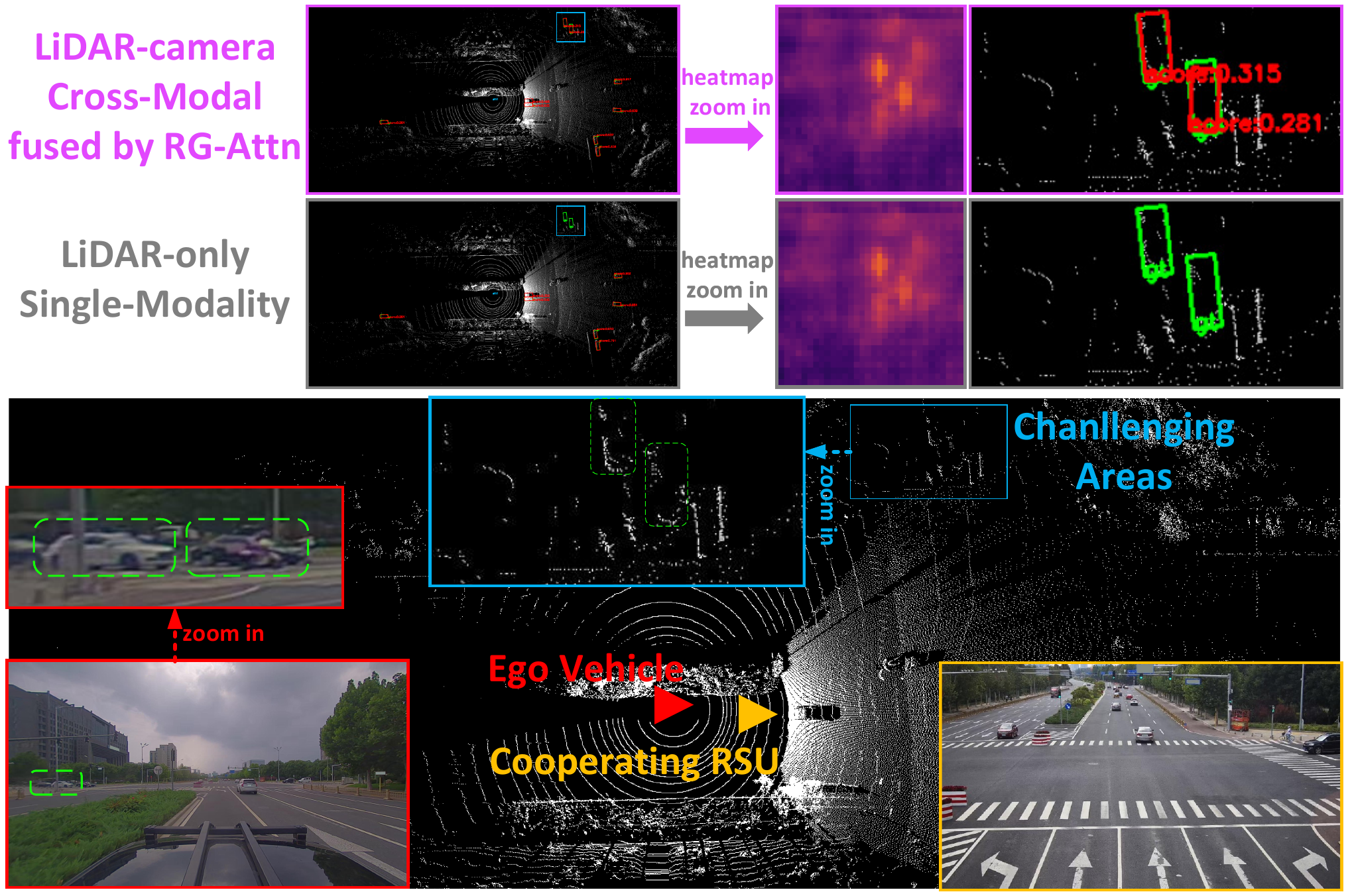}}
\caption{A representative scenario where the aggregated LiDAR BEV features from the ego and cooperating agents are insufficient to detect \textcolor{cyan}{challenging regions}. By fusing camera features—particularly from the \textcolor{red}{ego agent} in this frame, where objects are clearly captured—into the BEV space, additional semantic cues are introduced, leading to improved detection. This enhancement is evident in both the heatmap (brighter and with higher contrast) and the final detection output.}
\label{fig0}
\end{figure}

\begin{figure*}[ht]
\centerline{\includegraphics[width=1.0\linewidth]{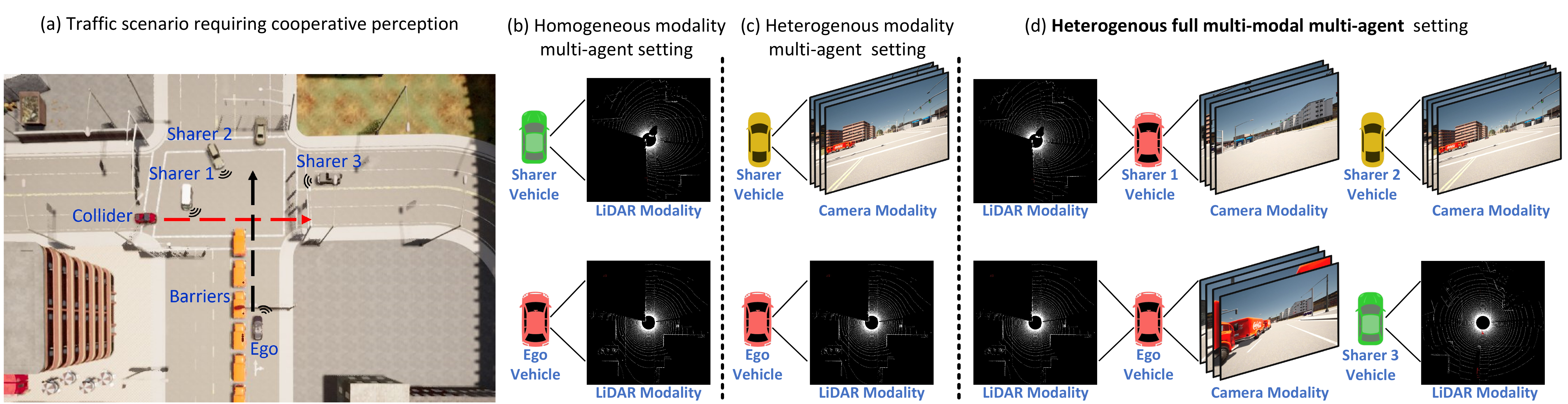}}
\caption{(a) A representative traffic scenario where cooperative perception enables agents to see through occlusions and prevent collisions. (b) Homogeneous modality multi-agent setting for cooperative perception. (c) Heterogeneous modality multi-agent setting with a restriction on single modality per agent for cooperative perception. (d) Heterogeneous full multi-modal multi-agent setting without restrictions on the number or types of modalities for cooperative perception, partially covered by PTP/PRGAF and fully covered by CoS-CoCo.}
\label{fig1}
\end{figure*}

The famous phrase \textit{``United we stand, divided we fall"} by Aesop aptly captures the essence of multi-agent cooperative perception. Shared and fused perception information serves as a crucial stepping stone—providing augmented environmental awareness as illustrated in \cref{fig0}—that enables more informed maneuvering decisions, helping to prevent traffic accidents as shown in \cref{fig1}(a). By overcoming the limitations of single-agent perception through the exchange of processed sensing data among multiple agents, challenges like non-line-of-sight (NLOS) occluded blind zones, partial object detection and limited detection range can be significantly mitigated. Typically, cooperative perception relies on the integration of V2X wireless communication, sensor data processing, and fusion modules to form a unified collaborative framework. Beyond intelligent transportation systems, cooperative perception also supports multi-robot use cases like factory automation and panoramic imaging.

While our focus is on cooperative perception, advances in single-agent perception continue to provide valuable foundations—such as stronger vision backbones and attention mechanisms for feature extraction and correlation. A persistent debate exists between single-modal and multi-modal designs in autonomous driving, especially concerning the trade-off between cost and performance. Despite deployment challenges, multi-modal approaches have shown empirical advantages in accuracy, robustness, and detection range, consistently outperforming single-modal baselines across public benchmarks.

In contrast, most cooperative perception frameworks rely on fusing a single sensor modality (e.g., LiDAR or camera) across multiple agents, as illustrated in \cref{fig1}(b). Recent methods like HM-ViT \cite{hmvit} and HEAL \cite{heal} have partially enabled heterogeneous setups, where each agent contributes only one modality (see \cref{fig1}(c)). However, fully multi-modal, multi-agent cooperation shown in \cref{fig1}(d) remains largely unexplored. In fact, naively combining modalities per agent within these frameworks often leads to degraded performance, mainly due to the unreliability of camera-derived depth. This underscores a critical gap between current cooperative frameworks and the potential of full multi-modal multi-agent collaboration. Furthermore, as we aim to bridge this gap, it is equally important to ensure compatibility with agents equipped with only single sensing modality. This raises a central research question: \textit{How can we fully leverage every available sensor on every participating agent for cooperative perception?} While accuracy is the primary goal, practical deployment also demands efficiency, communication feasibility, and system-level scalability.

In this paper, we propose Radian-Glue Attention (RG-Attn), a lightweight and effective module for multi-modal feature fusion. RG-Attn samples the LiDAR-derived Bird’s Eye View (BEV) feature map using radian divisions aligned with each camera's field of view (FOV), enabling column-wise attention-based fusion. This allows semantic enrichment of BEV features via projected camera cues while maintaining high computational efficiency. Building on RG-Attn, we introduce three cooperative perception architectures: Paint-To-Puzzle (PTP), Co-Sketching-Co-Coloring (CoS-CoCo) and Pyramid-RG-Attn Fusion (PRGAF), designed to address diverse deployment needs. PTP performs cross-modal fusion within each agent before a single-stage inter-agent fusion. It assumes LiDAR-equipped agents and unifies sharing data format. CoS-CoCo adopts a two-stage inter-agent fusion process: first fusing LiDAR BEV features among LiDAR-equipped agents, followed by camera feature enhancement from camera-equipped agents. This structure supports heterogeneous configurations, allowing participation from LiDAR-only, camera-only, or multi-modal agents. PRGAF integrates RG-Attn directly into the multi-scale pyramid structure of cross-agent fusion, fully leveraging camera semantics at multiple resolutions to enrich LiDAR features—albeit with significantly higher computational cost. These designs reflect distinct trade-offs based on the aforementioned motivations: PTP favors efficiency, CoS-CoCo balances generality and robustness, and PRGAF targets peak detection performance. All three outperform existing approaches by exploiting available sensor modalities while supporting real-time inference. In summary, our main contributions are:

\begin{itemize}
\item We propose RG-Attn, a novel and generalizable cross-modal fusion module. RG-Attn supports both intra-agent and inter-agent cross-modal fusion, delivering robust performance with high computational efficiency.

\item We design three cooperative architectures that integrate cross-modal and cross-agent fusion, each tailored to different deployment scenarios and performance trade-offs.

\item Extensive experiments on cooperative perception benchmarks demonstrate the effectiveness of our fusion module and architectures, achieving state-of-the-art performance.
\end{itemize}

%% file: sec/2_relatedworks.tex
\section{Related Works}
\label{sec:relatedworks}

\subsection{Single-Agent Perception}
In single-agent perception, a wide range of single-modal approaches have laid the groundwork for both single-modal and multi-modal research. Benchmarks such as KITTI and nuScenes \cite{nuscenes} have driven advances in both the LiDAR and camera tracks. For LiDAR, methods like PointNet \cite{pointnet}, pillar encoding, and voxel encoding have significantly improved feature aggregation, enabling compact and accurate scene representations. In the camera domain, performance has rapidly advanced from monocular setups to multi-view inputs and depth-aware methods, notably with the introduction of Lift-Splat-Shoot (LSS) \cite{LSS}. Transformer-based architectures like ViT and Point Transformer \cite{vit, PointTransformer} have further boosted both LiDAR and camera pipelines. Recently, these two modalities have increasingly converged under BEV-based frameworks \cite{focalformer3d, temporal}, which unify 3D object queries and positional encoding across modalities.

Multi-modal approaches, in contrast, focus on bridging LiDAR and camera data. Early work such as PointPainting \cite{pointpainting} injected camera semantics into LiDAR point clouds. Later, DETR-style models \cite{detr3d} leveraged attention to model cross-modality relationships. The rise of BEV-based fusion methods \cite{BEVFusion, BEVFormer} further enabled cross-modality alignment via attention, benefiting from BEV’s unified spatial representation and query efficiency. A more recent trend \cite{sparselif, las} challenges the reliability of depth estimated from camera data. Instead of relying on depth, these methods directly project camera features into LiDAR-derived BEV space using attention mechanisms, achieving more robust fusion.

\subsection{Multi-Agent Cooperative Perception}
Multi-agent cooperative perception offers distinct advantages by approaching perception tasks from a broader, system-level perspective. With the emergence of rich datasets \cite{dair, opv2v, v2v4real, tum}, the field has undergone multiple waves of technical innovation. Early efforts explored early fusion of raw sensor data \cite{early} to retain signal fidelity and late fusion of detection results \cite{late} to reduce communication overhead. More recent approaches \cite{cobevt, bm2cp, hmvit, heal} adopt intermediate feature fusion to trade off perception accuracy and V2X bandwidth, achieving superior performance in LiDAR-centered settings. Meanwhile, camera-centered works \cite{coocc, coverl} demonstrate the value of dense semantics.

Various auxiliary advances have strengthened the cooperative framework: Who2Com and Where2comm \cite{who2com, where2comm} prioritized selective message transmission, FedBEVT \cite{fedbevt} introduced federated learning to preserve privacy, and Coopernaut and ICOP \cite{coopernaut, icop} demonstrated performance gains in end-to-end autonomous driving. For practical real-world deployment, studies such as CoAlign \cite{coalign} and CBM \cite{cbm} have demonstrated significant advancements in minimizing relative localization errors—essential for reliable coordinate transformations. Additionally, research in vehicular communication \cite{v2x1, v2x2, v2x3} has enhanced protocol efficiency to better support perception-layer demands. Fusion techniques have evolved from simple concatenation to transformer-based and pyramid-based designs. However, most prior work remains limited to single-modality. Recent advances such as HM-ViT \cite{hmvit}, BM2CP \cite{bm2cp}, and HEAL \cite{heal} have begun incorporating multi-modality. Yet, HM-ViT and HEAL only allow one shared modality per agent, and although BM2CP supports multi-modality per agent, its performance still lags behind the LiDAR-only HEAL. The existing gap is how much further can perception be improved if all available multi-agent sensor sources are efficiently fused, this paper aims to address this critical gap.

%% file: sec/3_methodology.tex
\section{Methodology}
\label{sec:methodology}

\subsection{Radian-Glue-Attention (RG-Attn)}
Since the direct yet unreliable depth estimation originating from camera data is discarded, the key to effective cross-modal fusion lies in accurately projecting 2D semantic features from camera views onto the BEV feature map generated by the LiDAR backbone. Let the solid LiDAR BEV feature map be denoted as $F^{\text{bev}}_{j} \in \mathbb{R}^{C_1 \times H_1 \times W_1}$ and the camera feature from camera $k$ on agent $i$ be $F^{\text{cam}}_{ik} \in \mathbb{R}^{C_2 \times H_2 \times W_2}$, where $j$ and $i$ represent agent indices and $ik$ identify the camera $k$ mounted on agent $i$. To enable cross-modal fusion at both intra-agent and inter-agent levels, we compute the transformation matrix $T_{i \rightarrow j}$ from agent $i$ to agent $j$ in BEV space, and use the camera-to-agent transform $t_{ik\rightarrow i}$ to derive the camera's BEV coordinate location $(x, y)$ on agent $j$ as $t_{ik\rightarrow j} = T_{i \rightarrow j} \cdot t_{ik\rightarrow i}$. The rotation component $R_{i \rightarrow j}$ of $T_{i \rightarrow j}$ and the rotation matrix $R_{ik\rightarrow i}$ of camera $k$ to its mounted agent $i$ form $R_{ik\rightarrow j}$ as $R_{i \rightarrow j} \cdot R_{ik\rightarrow i}$, transforming the horizontal FOV range $\left[-\frac{\theta_{\text{FOV}}}{2}, \frac{\theta_{\text{FOV}}}{2}\right]$ from the local camera frame to the BEV frame of agent j $\left[\theta^{\text{start}}_{ik \rightarrow j}, \theta^{\text{end}}_{ik \rightarrow j} \right]$ as $\left[\text{atan}\left(R_{ik\rightarrow j} \cdot u\left(-\frac{\theta_{\text{FOV}}}{2}\right)\right), \text{atan}\left(R_{ik\rightarrow j} \cdot u\left(\frac{\theta_{\text{FOV}}}{2}\right)\right)\right]$, where $u$ denotes the unit direction vector.

\begin{figure}[!t]
\centerline{\includegraphics[width=\columnwidth]{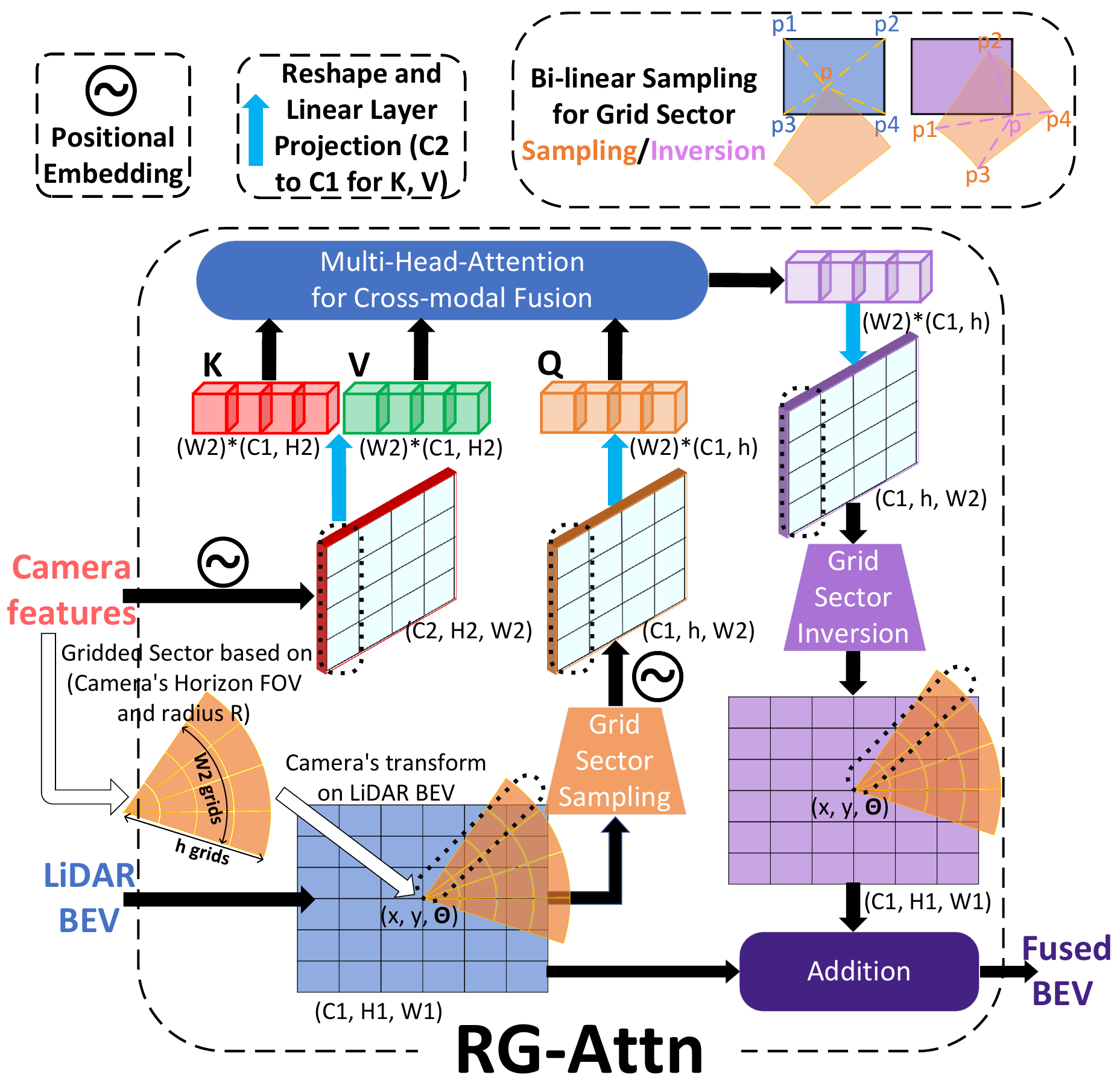}}
\caption{RG-Attn enables cross-modal fusion between LiDAR BEV and camera features. Camera parameters (FOV, range, extrinsics) define a polar grid sector for sampling a sub-BEV map from the LiDAR BEV. Both the sub-BEV and camera features are augmented with positional embeddings, from which queries (sub-BEV) and keys/values (camera) are generated via the PL process. Column-wise alignment along the width enables efficient multi-head attention. The fused features are then mapped back to the original BEV space via grid sector inversion and integrated through element-wise addition.}
\label{fig2}
\end{figure}

Once the relative transformation and horizontal FOV range of camera $k$ from agent $i$ are established on the target BEV map of agent $j$ as shown in \cref{fig2}, a geometric projection relationship is constructed, determining the angular span where the camera features will be projected. We discretize this angular span into $W_2$ sub-sectors, each aligned with a column in the camera feature matrix $F^{\text{cam}}_{ik} \in \mathbb{R}^{C_2 \times H_2 \times W_2}$. Each sub-sector is further radially divided into $h$ segments, forming a polar grid. The maximum projection radius $R$ is set to half the diagonal length of the BEV map to balance coverage and distortion. The number of radial segments $h$ to divide $R$, is set to match BEV height $H_1$ in our setting. Bilinear sampling is then used to project BEV features onto this polar grid, extracting a sampled sub-BEV map centered at $t_{ik\rightarrow j}$, with angular range $\left[\theta^{\text{start}}_{ik \rightarrow j}, \theta^{\text{end}}_{ik \rightarrow j} \right]$, angular resolution $W_2$, radial extent $R$ and radial resolution $h$. This defines a grid sector sampling configuration ${Set}_{ik \rightarrow j}$ as: $(t_{ik\rightarrow j},\ \left[\theta^{\text{start}}_{ik \rightarrow j}, \theta^{\text{end}}_{ik \rightarrow j} \right],\ W_2,\ R,\ h)$. Using this, a sector-shaped region is sampled from the BEV map and rescaled into a rectangular tensor $F^{\text{sub-bev}}_{ik \rightarrow j} \in \mathbb{R}^{C_1 \times H_1 \times W_2}$ as:
\begin{equation}
{F^\text{sub-bev}_{ik \rightarrow j}} = \text{GridSectorSample}(F^{\text{bev}}_{j},\ {Set}_{ik \rightarrow j}).
\label{eq:1a}
\end{equation}

The sampled LiDAR sub-BEV $F^{\text{sub-bev}}_{ik \rightarrow j} \in \mathbb{R}^{C_1 \times H_1 \times W_2}$ aligns with the camera feature map $F^{\text{cam}}_{ik} \in \mathbb{R}^{C_2 \times H_2 \times W_2}$ on width dimension. This column-wise alignment enables fusion to be performed in $C \times H$ spaces per column, instead of the full $C \times H \times W$ BEV space, thereby reducing the complexity from quadratic to linear in $W$. 

Both features are first augmented with positional embeddings, with queries generated from the sub-BEV feature and keys/values from the camera feature via reshaping and linear layer projection, collectively denoted as the PL (Positional-embedding-reshape-Linear-layer-Projection) process. Multi-head attention is applied in parallel within the $C \times H$ space of each column across the $W$-dimension for cross-modal fusion:
\begin{equation}
{\overline{F}^\text{fus-bev}_{ik \rightarrow j}} = \text{Attn}(\text{PL}(F^\text{sub-bev}_{ik \rightarrow j}),\ \text{PL}(F^{\text{cam}}_{ik}),\ \text{PL}(F^{\text{cam}}_{ik})).
\label{eq:1b}
\end{equation}

The camera semantics are thus ``glued" onto the sub-BEV representation in a radian-aligned, column-wise manner. Subsequently, grid sector inversion utilizing the same geometric correspondence but reversing the bilinear sampling direction, is applied to inversely sample the enhanced but distorted feature back to the original BEV grid as:
\begin{equation}
{F^\text{fus-bev}_{ik \rightarrow j}} = \text{GridSectorInverse}(\overline{F}^\text{fus-bev}_{ik \rightarrow j},\ {Set}_{ik \rightarrow j}).
\label{eq:1c}
\end{equation}

Finally, the enhanced feature map $F^{\text{fus-bev}}_{ik \rightarrow j}$ is element-wise added to the original BEV feature map $F^{\text{bev}}_{j}$ to obtain the fused output $F^{\text{fus-bev}}_{j + ik} \in \mathbb{R}^{C_1 \times H_1 \times W_1}$ as:
\begin{equation}
{F^\text{fus-bev}_{j + ik}} = F^{\text{bev}}_{j} + F^\text{fus-bev}_{ik \rightarrow j}.
\label{eq:1d}
\end{equation}

The entire RG-Attn pipeline in \cref{fig2}—comprising grid sector sampling, attention-based fusion, inverse sampling, and feature integration—can be compactly expressed as:
\begin{equation}
{F^\text{fus-bev}_{j + ik}} = \text{RG-Attn}(F^{\text{bev}}_{j},\ F^{\text{cam}}_{ik}).
\label{eq:1}
\end{equation}

\begin{figure*}[ht]
\centerline{\includegraphics[width=1.0\linewidth]{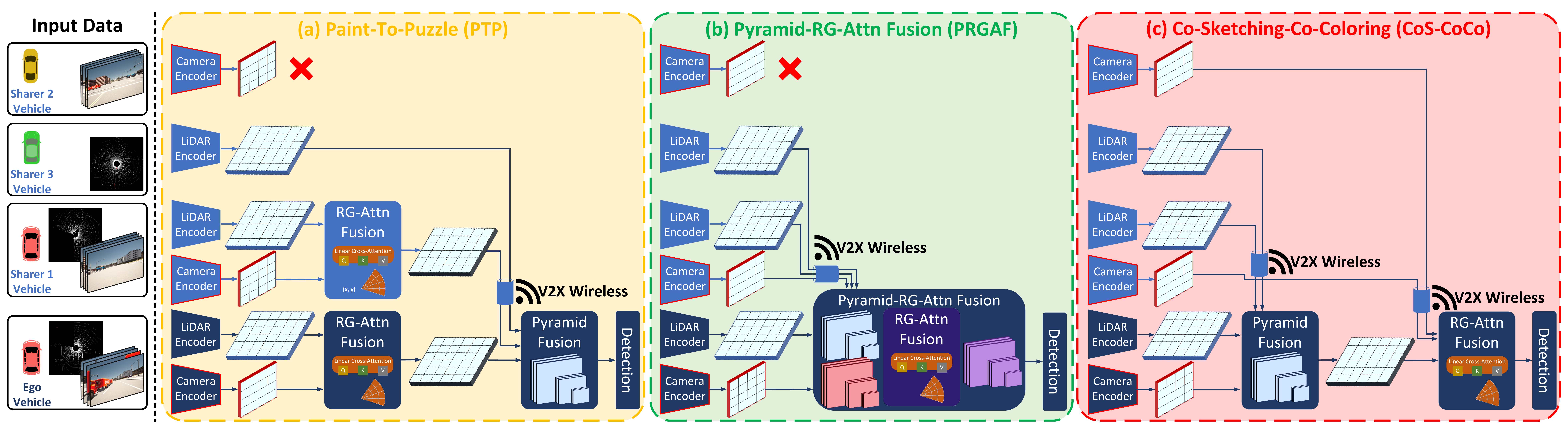}}
\caption{Illustration of the PTP, PRGAF and CoS-CoCo architectures. In all designs, all vehicles are capable of executing the perception pipeline independently, with or without cross-agent features. For illustration clarity, one vehicle (e.g., the ego) is shown as the receiver of external features and performs the complete cooperative perception process.}
\label{fig3}
\end{figure*}

\subsection{RG-Attn enabled Cooperative Perception}
\subsubsection{Paint-To-Puzzle (PTP)}

The core idea of PTP is that each agent constructs the cross-modal fused BEV feature map individually before engaging in cross-agent fusion. As illustrated in \cref{fig3}(a), agents equipped with both LiDAR and cameras first ``\textbf{paint}" their local environments and then ``\textbf{puzzle}" these together in a collaborative manner. Accordingly, the RG-Attn module is applied solely within each individual agent in PTP, generating $F^{\text{fus-bev}}_{ik \rightarrow i}$ multiple times, once for each camera $k$ onboard agent $i$. These semantically enriched features are subsequently aggregated to produce $F^{\text{fus-bev}}_{i + \sum_{k=1}^{n} ik}$ as:
\begin{equation}
F^{\text{fus-bev}}_{i + \sum_{k=1}^{n} ik} = \text{RG-Attn}(F^{\text{bev}}_{i},\ \{F^{\text{cam}}_{ik} \mid k = 1,2, \ldots,n \}),
\label{eq:2a}
\end{equation}
where $i \in AgentsSet_{LiDAR+camera}$. For collaborative LiDAR-only agents, the original BEV feature map is retained and directly used in the subsequent ``puzzle" step.

The Pyramid Fusion module from HEAL \cite{heal} is adopted as the backbone for the ``puzzle" part, fusing all available BEV feature maps into a richer global BEV space:
\begin{equation}
F_\text{PTP} = f_\text{pyramid fusion} \left( F^{\text{fus-bev}}_{i + \sum_{k=1}^{n} ik},\ F^{\text{bev}}_{m} \right),
\label{eq:2b}
\end{equation}
where $i \in AgentsSet_{LiDAR+camera}$ and $m \in AgentsSet_{LiDAR\_only}$. The multi-scale architecture and foreground-aware mechanisms of the fusion module enhance the integration of both semantically enriched and raw LiDAR features from diverse perspectives. Camera-only agents are excluded from this process due to their lack of reliable, depth-grounded BEV features. Importantly, the PTP design standardizes all shared perception into a unified BEV-based payload format.

\subsubsection{Co-Sketching-Co-Coloring (CoS-CoCo)}

As shown in \cref{fig3}(c), the CoS-CoCo framework is structured into two distinct fusion stages: \textbf{Co-Sketching}, which fuses LiDAR BEV features among LiDAR-equipped agents to construct a shared environmental ``skeleton,'' and \textbf{Co-Coloring}, which overlays semantic information from camera views onto this skeleton.

In the Co-Sketching stage, similar to the BEV fusion in PTP, the Pyramid Fusion module is adopted to aggregate all available LiDAR BEV features into a unified and robust BEV representation $F^{\text{bev}}_{pyr} \in \mathbb{R}^{C_1 \times H_1 \times W_1}$ as:
\begin{equation}
F^{\text{bev}}_{pyr} = f_\text{pyramid fusion} \left( F^{\text{bev}}_{l} \right),
\label{eq:3a}
\end{equation}
where $l \in AgentsSet_{LiDAR}$, indicating that each LiDAR-equipped agent jointly ``sketches'' the spatial foundation.

During the Co-Coloring stage, all camera feature maps—regardless of their source agents—are projected onto the shared BEV skeleton using the RG-Attn module, which provides robust cross-modality alignment in heterogeneous multi-agent settings. Based on the total number of camera-equipped agents and their respective cameras, the fusion process is expressed as:
\begin{equation}
F_\text{CoS-CoCo} = \text{RG-Attn}(F^{\text{bev}}_{pyr},\ \{F^{\text{cam}}_{ck} \mid k = 1,2, \ldots,n \}),
\label{eq:3b}
\end{equation}
where $c \in AgentsSet_{camera}$ and $n$ is the number of cameras per agent in the set.

A key advantage of CoS-CoCo lies in its ability to re-integrate camera-only agents into the collaborative perception pipeline by deferring camera-to-BEV projection to a centralized, skeleton-based stage. However, this approach requires managing two distinct formats of cooperative payloads—LiDAR BEV and camera 2D features.

\begin{figure}[!t]
\centerline{\includegraphics[width=\columnwidth]{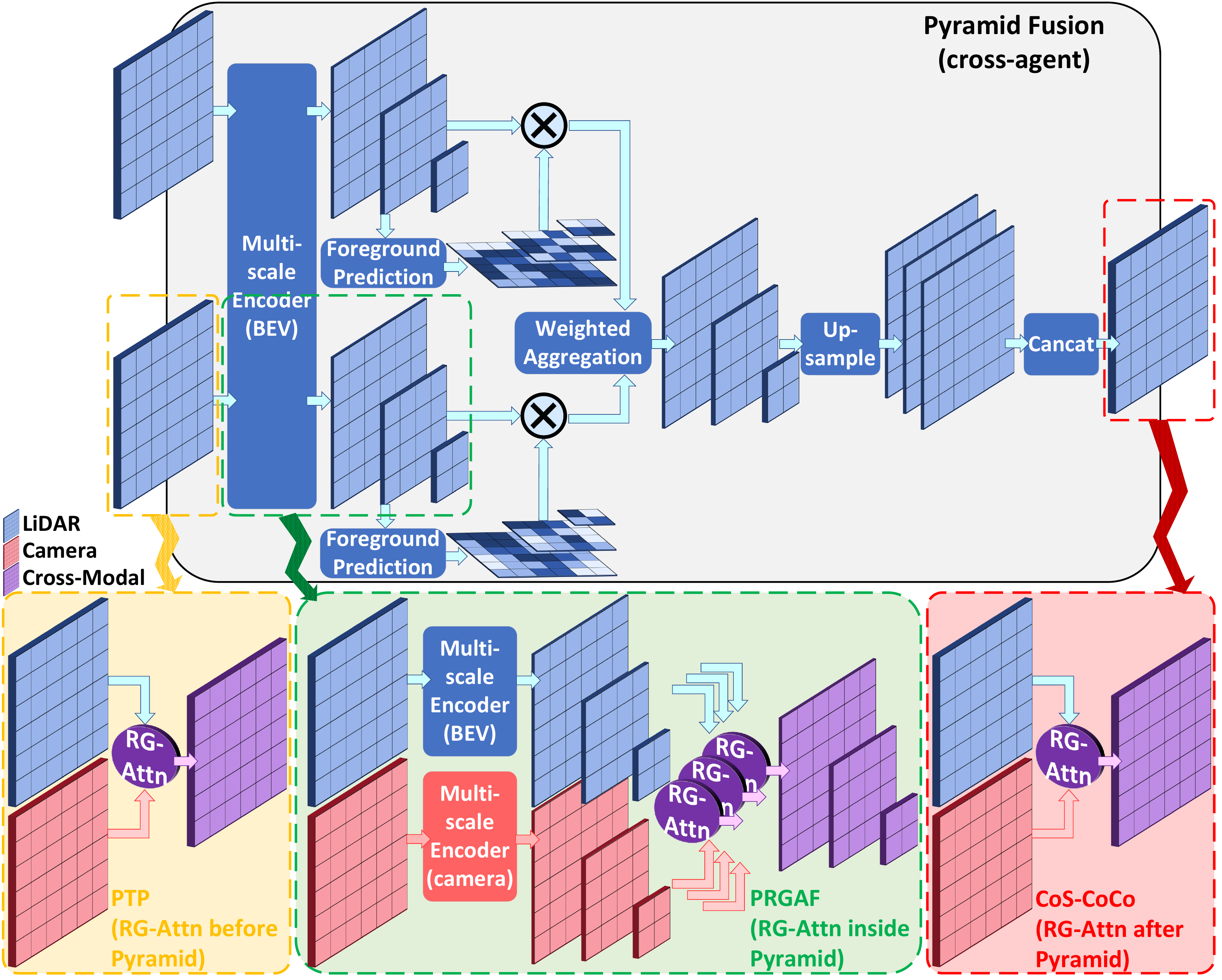}}
\caption{Structural comparison of PTP, PRGAF, and CoS-CoCo architectures, highlighting where and how RG-Attn is applied in relation to the cross-agent Pyramid Fusion module.}
\label{fig4}
\end{figure}

\subsubsection{Pyramid-RG-Attn Fusion (PRGAF)}

In contrast to the modular pipelines of PTP and CoS-CoCo, we design a performance-driven variant that integrates RG-Attn directly into the multi-scale pyramid fusion structure, as illustrated in \cref{fig3}(b). Unlike PTP and CoS-CoCo—where RG-Attn is applied either before or after the pyramid at a single scale (as shown in \cref{fig4})—this unified architecture performs RG-Attn at each resolution level within the pyramid, enabling hierarchical cross-modal enhancement prior to cross-agent fusion.

Concretely, we extract multi-scale BEV and camera features at three resolutions (e.g., widths \(W = 64, 128, 256\)), and apply RG-Attn independently at each level to fuse camera semantics into the BEV space per agent as \cref{eq:2a} to get $F^{\text{fus-bev}}_{i + \sum_{k=1}^{n} ik, (s)}$, where \(s\) indexes the resolution level. The resulting fused features are then passed to perform occupancy-aware alignment and weighted aggregation at each scale, followed by upsampling and concatenation to produce the final fused BEV representation. By embedding RG-Attn throughout the entire pyramid, the integrated architecture fully leverages multi-scale camera semantics to enrich BEV features and maximize detection performance.

%% file: sec/4_experiments.tex
\section{Experiments and Results}
\label{sec:experiments}

\subsection{Datasets}
To evaluate the effectiveness of our multi-modal multi-agent framework, we conduct experiments on the DAIR-V2X and OPV2V datasets. Our model design is primarily guided by the DAIR-V2X dataset, which presents real-world cooperative perception challenges with data collected in urban Beijing. DAIR-V2X contains 9K frames, each consisting of raw sensor data from a vehicle agent and a roadside unit (RSU), each equipped with a single camera and a LiDAR of differing specifications. In contrast, the OPV2V dataset, simulated in CARLA, includes over 11K frames across diverse traffic scenarios. Each OPV2V frame involves 2 to 7 collaborating vehicles, each uniformly equipped with one LiDAR and four cameras.

Across both datasets, we evaluate three agent configurations: LiDAR-only, camera-only, and LiDAR–camera both. For OPV2V, camera-equipped agents utilize all four 800×600 resolution cameras, while LiDAR-equipped agents use the default 64-beam 360° LiDAR. In DAIR-V2X, both vehicle and RSU provide 1920×1080 camera input; however, notable differences in camera height and intrinsic parameters exist between the two. The LiDAR configurations also differ: vehicles use a 40-beam 360° LiDAR, while RSUs employ a 300-beam LiDAR with a 100° FOV.

\subsection{Settings}

\textbf{Implementation details:} We adopt unified encoders for raw data processing: PointPillar \cite{pointpillar} for LiDAR point cloud and the first five sequential layers of ResNet101 for camera images. The LiDAR BEV map is down-sampled $2\times$ and further reduced to a shape of [64, 128, 256] using 3 consecutive ConvNeXt \cite{convxt} blocks, with a grid size of [0.4m, 0.4m]. The feature matrix for each camera sensor after the encoder is in shape [8, 144, 256], as we fix the target channel, width and height of the output to deal with the different resolution specification in two datasets. Multi-head attention is configured with corresponding learnable embeddings, attention heads of 8, and a dropout rate of 0.1. The pyramid fusion conducts multi-level fusion with widths (i.e., the last dimension of BEV map feature shape) at 256, 128 and 64 consequently. For fair comparison with existing approaches, the detection range in both training and evaluation is set to $x \in [-102.4m,\ +102.4m],\ y \in [-51.2m,\ +51.2m]$. A Non-Maximum Suppression (NMS)-based object detection head is added, which processes the outputs of the classification, regression, and orientation predictions to generate final detections. Average precision (AP) is then calculated at different intersection-over-union (IoU) thresholds.

\noindent \textbf{Training configurations:} Three loss functions—for classification, regression, and orientation—are used, with the foreground map incorporated into the loss calculation. We adopt the Adam optimizer with an initial learning rate of 0.002, which is reduced by a factor of 0.1 from epoch 15 to 25 for DAIR-V2X (30 epochs in total) and from epoch 35 to 40 for OPV2V (40 epochs in total). Training on a single NVIDIA RTX 6000 Ada takes approximately 6 hours for DAIR-V2X and 36 hours for OPV2V.

\subsection{Quantitative \& Visualization Results}

As shown in \cref{tab:tab1}, we compare our approach with the best performances of existing methods \cite{opv2v, disconet, v2xvit, cobevt, bm2cp, hmvit, heal}, each evaluated under its optimal modality configuration as reported in their original papers. For a fair and consistent comparison, we re-implemented or reproduced all listed results under identical experimental settings. The number of collaborating agents is fixed at 2 for DAIR-V2X and up to 5 for OPV2V, with the same detection range applied. A key distinction is that most compared methods (except BM2CP) achieve their best results using only LiDAR data from all agents, as documented in their papers. 

We also evaluate our methods under varying numbers and combinations of agents and modalities on both datasets, as shown in \cref{tab:tab2}, where ``+" separates two agents, ``L" denotes LiDAR-only, ``C" camera-only, and ``LC" LiDAR-camera-both. All models are trained only once with full multi-modal multi-agent setting (i,e,. LC+LC), and directly used for inference in all other configurations. The multi-modal results of HEAL and CoBEVT marked with ``*" in \cref{tab:tab2}, are achieved by their proposed BEV fusion modules to fuse LiDAR-BEVs and camera-BEVs (with estimated depth from camera data) from participating agents.

\begin{figure}[!t]
\centerline{\includegraphics[width=\columnwidth]{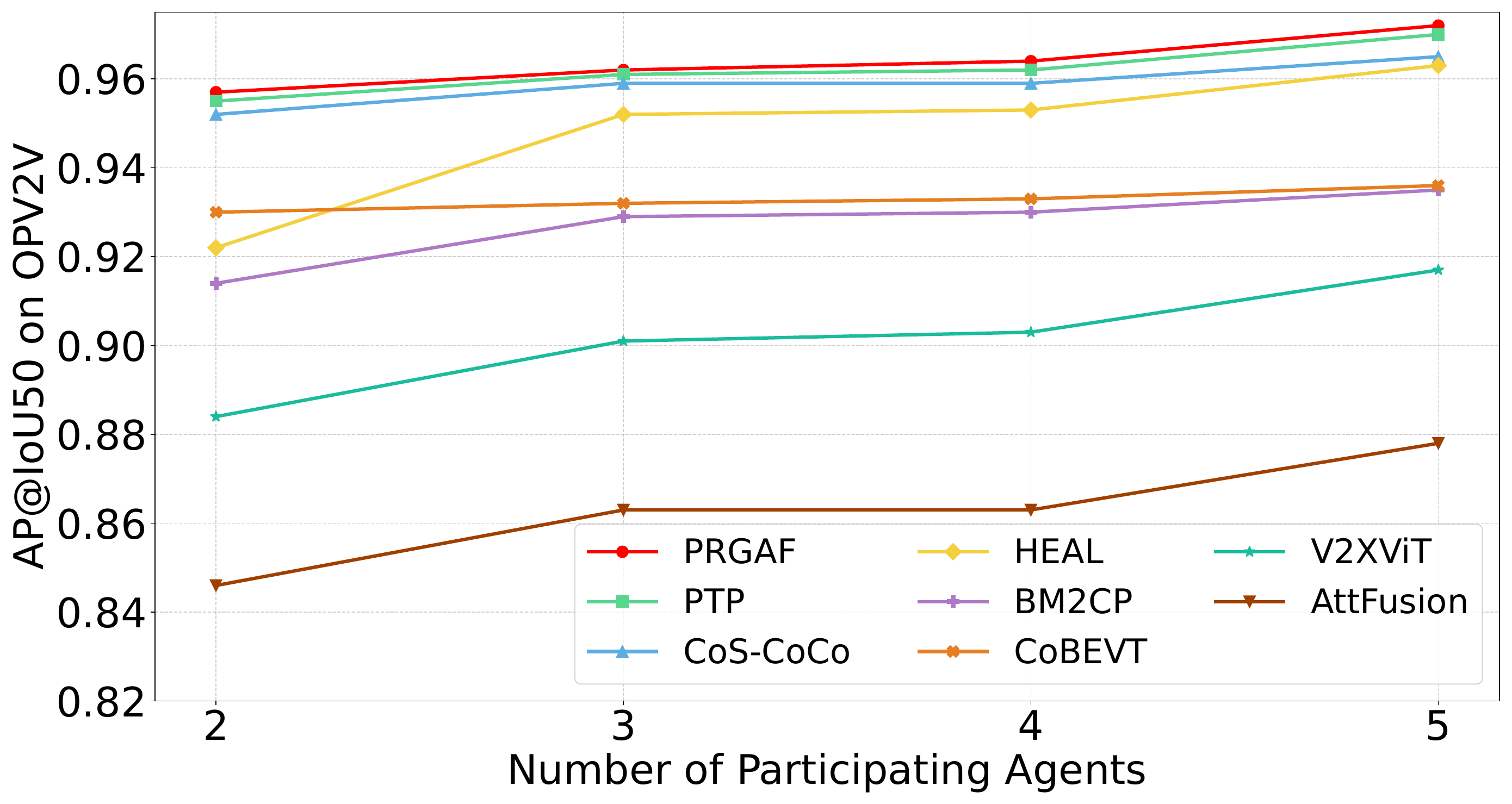}}
\caption{Comparison of AP50 scores on OPV2V with different maximum agent counts across our methods and key baselines.}
\label{fig5}
\end{figure}

Notably, CoBEVT and HEAL show performance drops when BEVs generated from camera data are included, as seen in the AP30 column of \cref{tab:tab1} and the LC+LC column under DAIR-V2X (AP30) in \cref{tab:tab2}, with decreases of 5.6\% and 19.9\%, respectively. A similar degradation appears on OPV2V, where HEAL and CoBEVT drop by 6.8\% and 27.1\% in AP50, as shown in \cref{fig5} and the right side of \cref{tab:tab2}. In contrast, our approach fully exploits the complementary strengths of camera data by fusing its semantics directly into the robust LiDAR-BEV, rather than generating a separate camera-BEV. This advantage is especially clear when comparing HEAL with our method, as our cross-modal fusion extends HEAL’s LiDAR-only backbone. Moreover, our framework significantly outperforms BM2CP—another method designed for multi-modal input—highlighting the strength of our design.

\begin{table}[h]
\centering
\begin{tabular}{c|c|c|c|c|c} 
\hline
\multicolumn{2}{c|}{Dataset} & \multicolumn{2}{c|}{DAIR-V2X} & \multicolumn{2}{c}{OPV2V} \\
\hline
Method & Modal & AP30 & AP50 & AP50 &AP70\\
\hline
AttFusion \cite{opv2v} & L & 0.738 & 0.673 & 0.878 & 0.751\\

DiscoNet \cite{disconet} & L & 0.746 & 0.685 & 0.882 & 0.737\\

V2XViT \cite{v2xvit} & L & 0.785 & 0.521 & 0.917 & 0.790\\

CoBEVT \cite{cobevt} & L & 0.787 & 0.692 & 0.935 & 0.851\\

BM2CP \cite{bm2cp} & LC & 0.802 & 0.743 & 0.935 & 0.896\\
 
HM-ViT \cite{hmvit} & L & 0.818 & 0.761 & 0.950 & 0.873\\

HEAL \cite{heal} & L & 0.832 & 0.790 & 0.963 & 0.926\\
\hline
\rowcolor{red!5}
CoS-CoCo & LC & 0.854 & 0.811 & 0.965 & 0.937\\
\rowcolor{red!10}
PTP & LC & \textbf{0.862} & \textbf{0.817} & \textbf{0.970} & \textbf{0.945}\\
\rowcolor{red!15}
PRGAF & LC & \textbf{0.869} & \textbf{0.823} & \textbf{0.972} & \textbf{0.946}\\
\hline
\end{tabular}
\caption{Best performances of existing cooperative perception methods and our proposed approaches across different datasets, with identical modality setup per agent in each method.}
\label{tab:tab1}
\end{table}

\begin{table}[h]
\centering
\resizebox{\columnwidth}{!}{%
\begin{tabular}{c|c|c|c|c|c|c|c|c} 
\hline
Dataset & \multicolumn{4}{|c|}{DAIR-V2X (AP30)} & \multicolumn{4}{c}{OPV2V (AP50)}\\
\hline
Modality & LC & LC+C & LC+L & LC+LC & LC & LC+C & LC+L & LC+LC\\
\hline
CoBEVT* \cite{cobevt} & 0.146 & 0.553 & 0.589 & 0.588 & 0.472 & 0.604 & 0.647 & 0.643\\

HEAL* \cite{heal} & 0.237 & 0.574 & 0.692 & 0.776 & 0.581 & 0.636 & 0.733 & 0.854\\

BM2CP \cite{bm2cp} & 0.639 & 0.645 & 0.793 & 0.802 & 0.679 & 0.687 & 0.899 & 0.914\\

PTP & 0.707 & / & 0.743 & 0.862 & 0.820 & / & 0.875 & 0.955\\

PRGAF & 0.711 & / & 0.842 & 0.869 & 0.825 & / & 0.941 & 0.957\\
\rowcolor{green!15}
CoS-CoCo & 0.705 & \textbf{0.712} & \textbf{0.848} & 0.854 & 0.821 & \textbf{0.837} & \textbf{0.946} & 0.952\\
\hline
\end{tabular}
}
\caption{The performance comparison regarding the combination of agents number and modality setting in different approaches.}
\label{tab:tab2}
\end{table}

\begin{figure}[!t]
\centerline{\includegraphics[width=\columnwidth]{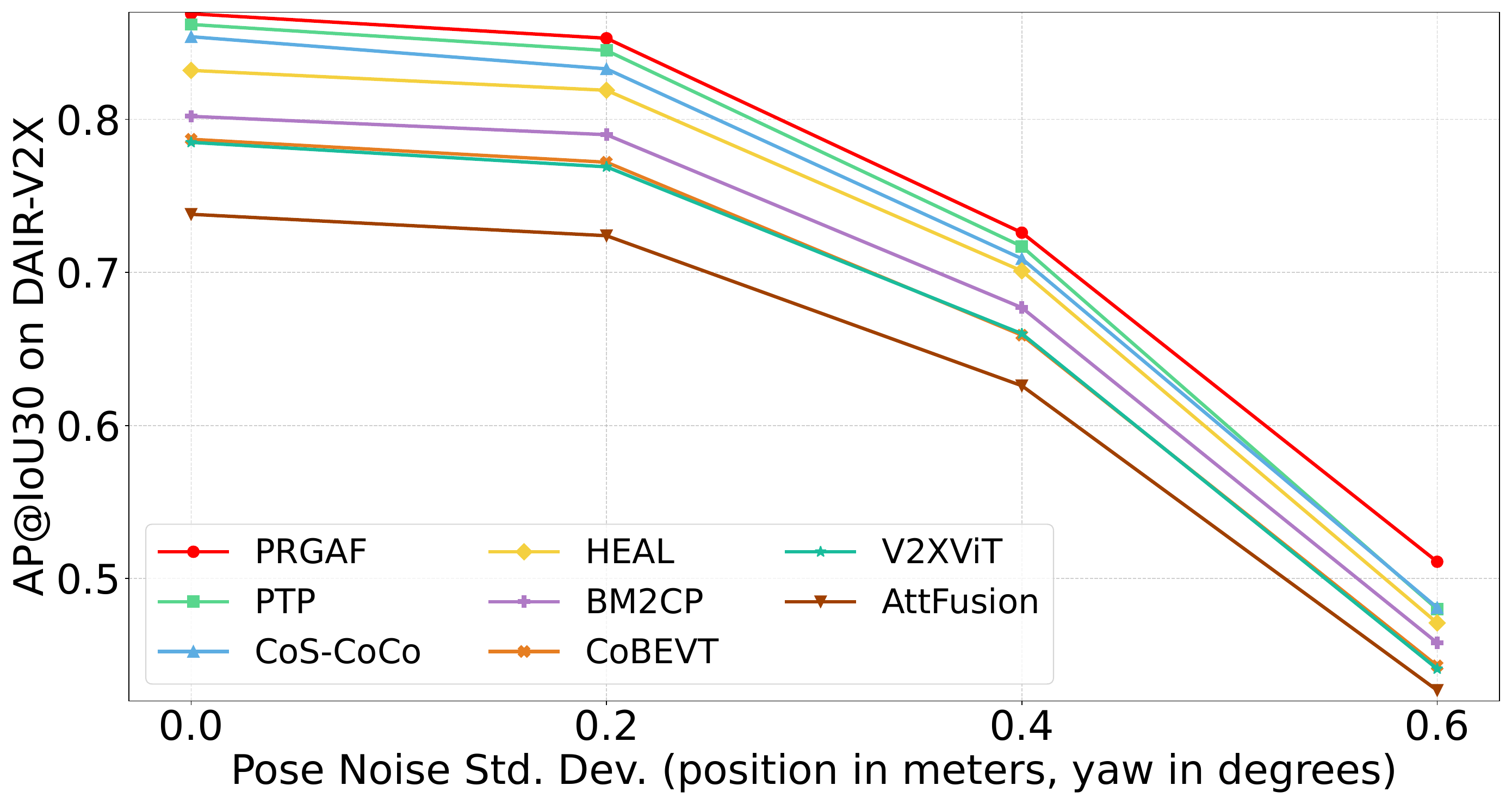}}
\caption{Comparison of AP30 scores on DAIR-V2X with pose noise: $\mathcal{N}(0, \sigma_p^2)$ for $x,y$ and $\mathcal{N}(0, \sigma_r^2)$ for yaw angle.}
\label{fig6}
\end{figure}

\begin{figure}[!t]
  \centering

  \begin{subfigure}{\linewidth}
    \centering
    \includegraphics[width=\columnwidth]{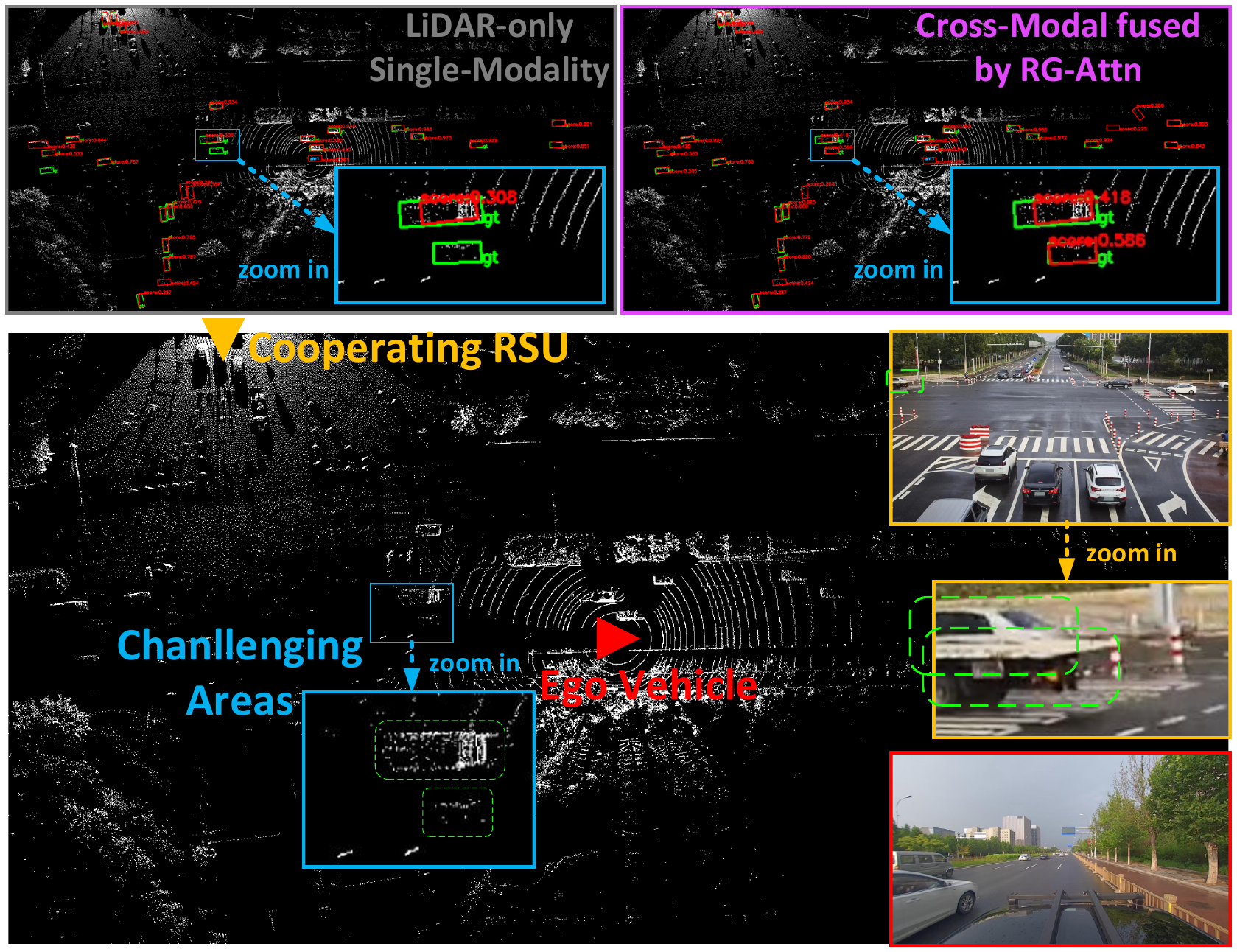}
    \caption{RG-Attn enhances BEV features for improved detection}
  \end{subfigure}

  \begin{subfigure}{\linewidth}
    \centering
    \includegraphics[width=\columnwidth]{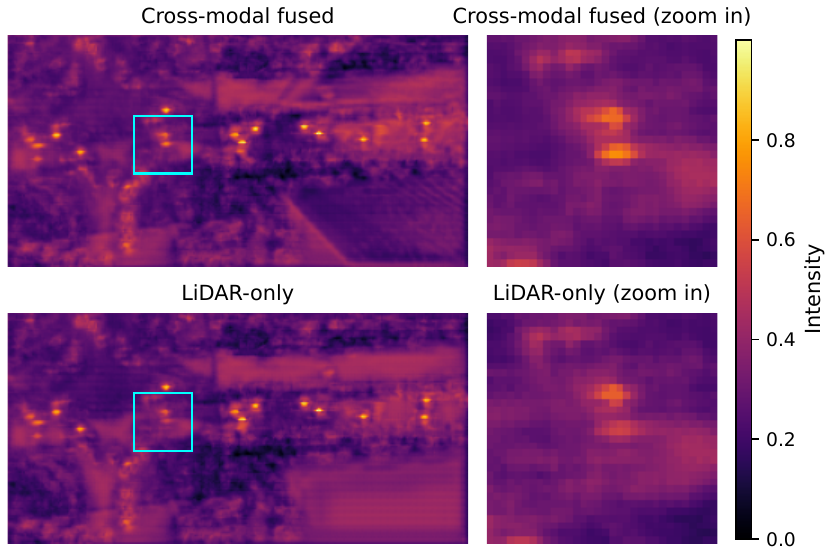}
    \caption{BEV feature heatmap – Channel 2/6 (classification head output)}
  \end{subfigure}

  \begin{subfigure}{\linewidth}
    \centering
    \includegraphics[width=\columnwidth]{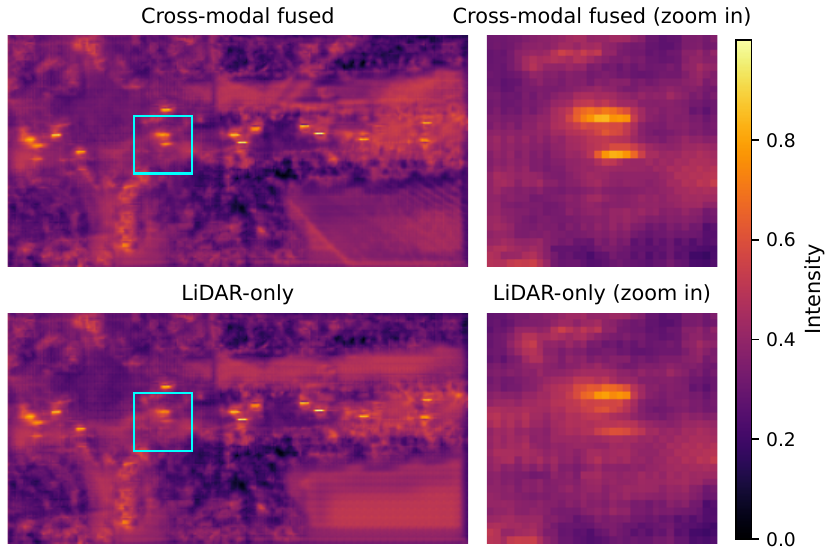}
    \caption{BEV feature heatmap – Channel 4/6 (classification head output)}
  \end{subfigure}

  \caption{Visualization of RG-Attn's effectiveness in a representative scenario.}
  \label{fig7}
\end{figure}

To assess the impact of the number of participating agents in cooperative perception on our proposed methods, we conduct a series of controlled experiments on the OPV2V dataset. As illustrated in \cref{fig5}, these experiments systematically evaluate how performance scales with varying numbers of collaborating agents, providing insights into the effectiveness and adaptability of our approach under different cooperation levels.

Additionally, we evaluate the impact of pose errors, as shown in \cref{fig6}, by adding Gaussian noise to the otherwise calibrated pose data (position and rotation) of DAIR-V2X, which is crucial for computing the transformation matrix that aligns features. Results show that our methods consistently outperform others under varying levels of pose noise.

The communication budget for transmitting intermediate BEV feature in PTP is 4 MB per agent per frame (uncompressed), but can be reduced to under 2 MB on average (47.56\% compression) using lossless methods like zlib, thanks to BEV feature sparsity. CoS-CoCo introduces a flexible load by adding $\sim$0.2 MB of uncompressed camera data per agent to the base 4 MB LiDAR-BEV, supporting LiDAR-only (4 MB), camera-only (0.2 MB), or both modalities (4.2 MB, same as PRGAF) to adapt to different bandwidth constraints. To reduce this substantial $\sim$4 MB per-frame budget, an autoencoder can compress features by shrinking channel dimensions for all three structures; tests show that a 32-fold compression (e.g., down to 0.125 MB for PTP) results in at most a 0.5\% AP30 drop on DAIR-V2X, with under 2 ms extra computation. The proposed RG-Attn module introduces minimal latency, requiring less than 4 ms per cross-modal fusion. The total inference time—from raw input to final fused BEV—is approximately 40 ms for PTP and CoS-CoCo, and 65 ms for PRGAF (measured in a two-agent cooperative setting).

In addition to quantitative results, we provide visualizations in \cref{fig7} to demonstrate the effectiveness of RG-Attn. Specifically, \cref{fig7}(a) illustrates a representative scenario where cross-modal fusion helps resolve challenging regions—areas where LiDAR-only BEV fails to support correct detection, but RG-Attn enhanced BEV succeeds. \cref{fig7}(b–c) further compare heatmaps from selected BEV feature map channels (extracted from the classification head and used as input to the NMS-based detection head) between the RG-Attn fused and LiDAR-only settings.

\subsection{Performance Analysis}

\textbf{Fusion Effectiveness:} As shown in \cref{tab:tab1}, all three RG-Attn-enabled architectures outperform existing methods, benefiting from the module’s effective cross-modal fusion capabilities. PRGAF achieves the highest accuracy on both benchmarks, surpassing the previous SOTA method HEAL by +3.7\% AP30 on DAIR-V2X and +2.0\% AP70 on OPV2V, by fully exploiting the potential of RG-Attn in a pyramid manner. This performance advantage remains consistent across varying numbers of participating agents and levels of pose noise, as illustrated in \cref{fig5} and \cref{fig6}. The gain stems from the intra-agent fusion stage, where camera semantics are effectively ``glued” onto the intact, sampled LiDAR-derived BEV map—yielding enriched and coherent representations. In contrast, during inter-agent fusion in CoS-CoCo, aligning the BEV map with a remote agent’s camera FOV can produce edge cases where the projected area reaches the LiDAR BEV boundary or lacks valid overlap, leading to fragmented, distorted, or missing semantics and reduced fusion effectiveness.

\noindent \textbf{Generalization:} As shown in \cref{tab:tab2}, CoS-CoCo demonstrates superior robustness and adaptability under heterogeneous sensor settings. Without any additional training or fine-tuning, it maintains competitive performance when integrating agents with different modality configurations, highlighting its suitability for real-world deployment.

\noindent \textbf{Ablation Component:} We evaluate different positional encoding strategies within RG-Attn in \cref{tab:tab3}. Removing positional encoding yields the poorest performance, though it still outperforms the no-cross-modal baseline. In contrast, both learnable encoding and depth-height hybrid encoding achieve comparable results under the PTP setting. In the hybrid scheme, the LiDAR BEV positional encoding corresponds to the radial depth of the sampled grid (via the GridSectorSample process), while the camera-side positional encoding reflects the vertical position (height) of semantics within the camera’s 2D feature column. The learnable encoding consists of two shared tensors, independently applied to the LiDAR BEV and camera feature columns prior to the column-wise attention operation.

\begin{table}[h]
\centering
\begin{tabular}{c|c|c|c} 
\hline
\multicolumn{1}{c|}{Dataset} & \multicolumn{3}{c}{DAIR-V2X}\\
\hline
Method & AP30 & AP50 &AP70\\
\hline
No Cross-Modal attention & 0.832 & 0.790 & 0.624\\

No positional encoding & 0.855 & 0.812 & 0.634\\

Depth-height hybrid encoding & 0.861 & 0.817 & 0.645\\

Learnable positional encoding & 0.862 & 0.823 & 0.642\\
\hline
\end{tabular}
\caption{Comparison of different positional encodings and a no cross-modal baseline for RG-Attn in the PTP architecture setting.}
\label{tab:tab3}
\end{table}

%% file: sec/5_conclusion.tex
\section{Conclusion and Future Work}
\label{sec:conclusion}

This work marks an initial step toward comprehensive multi-modal multi-agent fusion for cooperative perception, with opportunities for future improvements. The RG-Attn module can be further refined for improved cross-modal alignment under noisy sensing, and the framework could integrate the complementary strengths of all three structures to balance flexibility, accuracy, and efficiency. In summary, we demonstrate that our designs achieve high perception accuracy and computational efficiency in cooperative settings. We hope this work fosters continued exploration and discussion in the field of multi-modal multi-agent cooperative perception.